\pdfoutput=1

\documentclass[runningheads]{llncs}

\usepackage{latexsym}
\usepackage{multirow}
\usepackage{graphicx}
\usepackage{array}
\usepackage{tabularx}
\usepackage{lipsum}
\usepackage{microtype}  % This often helps with margin issues by optimizing character spacing.

\usepackage{arydshln}
\usepackage{rotating}
\usepackage{paralist}
\usepackage{booktabs}  % to add more space in the table
\usepackage{times}
\usepackage{latexsym}
\usepackage{marvosym} % Added for the \Letter symbol

\usepackage{url}

\usepackage[T1]{fontenc}
\usepackage[utf8]{inputenc}

\usepackage{inconsolata}

\usepackage{color}
\usepackage{hyperref}

\title{Generating patient cohorts from electronic health records using two-step retrieval-augmented text-to-SQL generation}

\author{Angelo Ziletti(\Letter)\orcidID{0000-0002-2978-6305} \and
Leonardo D’Ambrosi}
\authorrunning{A. Ziletti and L. D’Ambrosi}
\institute{Bayer AG \\
\email{\{angelo.ziletti, leonardo.dambrosi\}@bayer.com}}

\begin{document}
\titlerunning{RAG-based Text-to-SQL for Patient Cohorts}
\maketitle
\begin{abstract}
Clinical cohort definition is crucial for patient recruitment and observational studies, yet translating inclusion/exclusion criteria into SQL queries remains challenging and manual. We present an automated system utilizing large language models that combines criteria parsing, two-level retrieval augmented generation with specialized knowledge bases, medical concept standardization, and SQL generation to retrieve patient cohorts with patient funnels. The top-performing configuration achieves 0.75 F1-score in cohort identification on EHR data, effectively capturing complex temporal and logical relationships. These results demonstrate the feasibility of automated cohort generation for epidemiological research.
\keywords{Text-to-SQL \and Large Language Models \and Electronic Health Records \and Cohort Generation \and Retrieval-Augmented Generation.}
\end{abstract}

\section{Introduction} \label{sec:Intro}
Electronic Health Records (EHR) have become essential resources for clinical research, serving as a foundational component across multiple healthcare applications. In clinical trials, accurate cohort identification directly impacts patient recruitment success, study validity, timeline adherence, and cost management~\cite{sherman-et-al-2016-real}. For observational studies and epidemiological research, well-defined cohorts enable researchers to investigate disease patterns, treatment effectiveness, and health outcomes across specific patient populations, informing critical healthcare decisions from drug development to regulatory submissions~\cite{casey-et-al-2016-using,fda2024realworld}

\subsubsection{Problem Statement.} 
Converting clinical inclusion/exclusion criteria into accurate database queries requires addressing multiple technical challenges simultaneously: (1) mapping natural language criteria to precise computational logic while leveraging medical domain knowledge, (2) preserving complex temporal and logical relationships across multiple criteria, (3) mapping diverse medical concepts to standardized codes, (4) generating queries that conform to established data models, (5) returning patient-level identifiers with associated index dates to enable subsequent time-anchored analyses.

\subsubsection{Contributions.} 
This paper presents an approach for automatically generating patient cohorts from EHRs. 
Our main contributions are:

\begin{itemize}
\item A manually curated dataset of inclusion/exclusion criteria paired with SQL queries that conform to the Observational Medical Outcomes Partnership Common Data Model (OMOP-CDM)~\cite{omop-cdm-2023}. 
\item A two-level retrieval-augmented generation (RAG) framework that substantially outperforms simple prompting across all tested models, demonstrating robust performance in translating complex clinical criteria into executable SQL queries.
\item Open-source release of dataset, source code, and prompt configurations\footnote{\url{https://github.com/Bayer-Group/epi-cohort-text2sql-ecai2025}}.
\end{itemize}

The system has been deployed and is currently being evaluated at Bayer by epidemiologists and data analysts.

\begin{figure*}[htb]
\centering
\includegraphics[width=\textwidth]{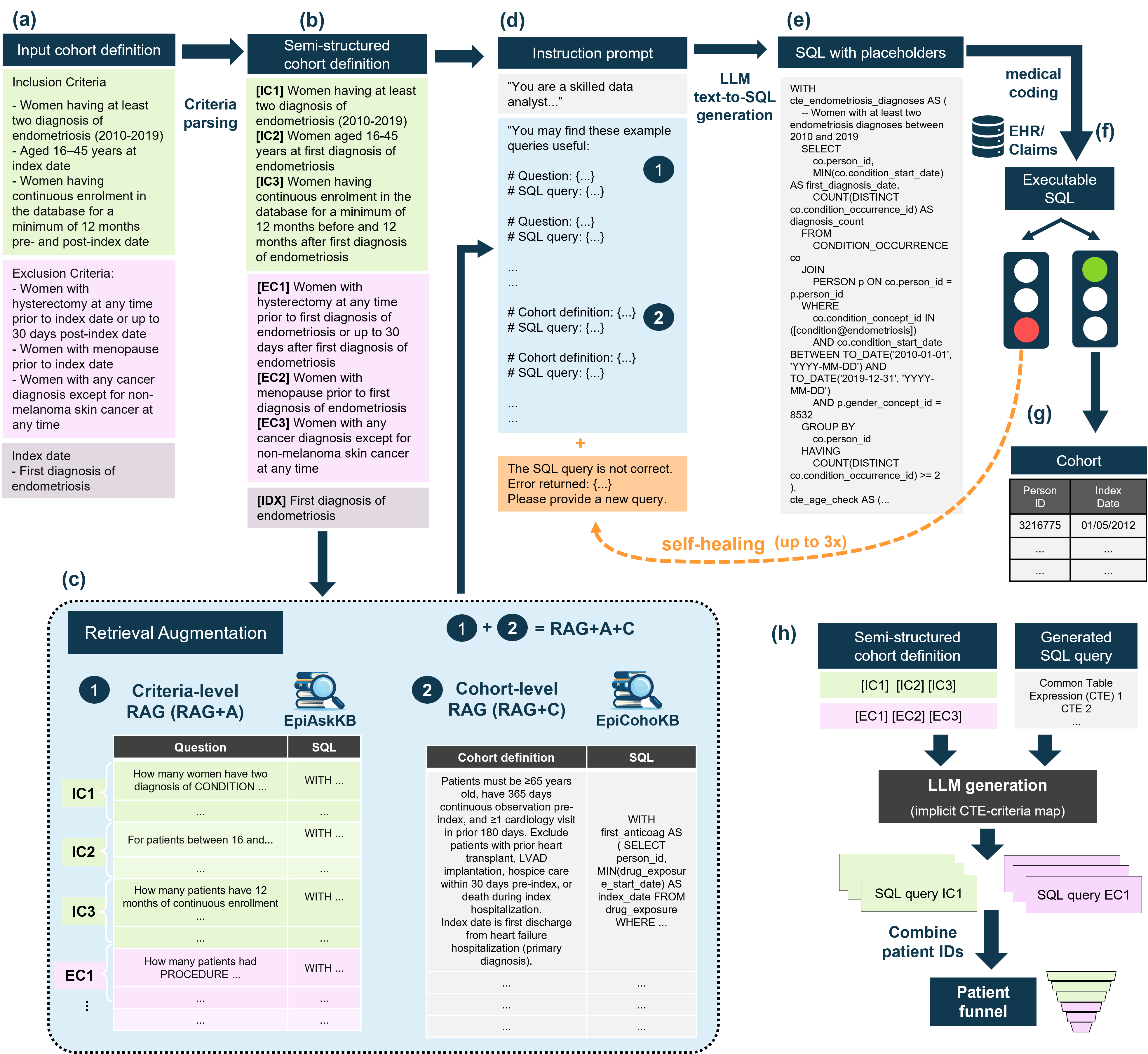}
\caption{(a-g) From inclusion/exclusion criteria in natural language to patient cohorts using electronic health record databases: end-to-end workflow. (h) Query decomposition and patient funnel generation through LLM-based processing.}
\label{fig:figure-workflow}
\end{figure*}

\section{Knowledge base generation} \label{sec:dataset}
\begin{table}[h]
\setlength{\tabcolsep}{4pt}
\renewcommand{\arraystretch}{1.2}  
\centering
\small
\begin{tabular}{l|c|c}
\hline
\hline
\textbf{Statistic} & \textbf{EpiAskKB} & \textbf{EpiCohoKB} \\
\hline
\multicolumn{3}{l}{\textit{Global Statistics}} \\
\hline
Number of samples & 115 & 108 \\
Number of different tables used & 13  & 12 \\
Number of different columns used & 40 & 32 \\
Unique medical concepts & 139 & 374 \\
\hline
\multicolumn{3}{l}{\textit{Text Characteristics}} \\
\hline
Question/criteria length [chars] & 82.2 $\pm$ 37.2 & 436.5 $\pm$ 82.1 \\
Question/criteria length [words] & 15.7 $\pm$ 7.6 & 76.3 $\pm$ 17.3 \\
Medical entities per query & 1.5 $\pm$ 1.0 & 8.0 $\pm$ 2.3 \\
\hline
\multicolumn{3}{l}{\textit{Criteria Complexity}} \\
\hline
\# of inclusion criteria & N/A & 4.1 $\pm$ 0.7 \\
\# of exclusion criteria & N/A & 3.7 $\pm$ 0.9 \\
\hline
\multicolumn{3}{l}{\textit{SQL Complexity}} \\
\hline
SQL length [chars] & 803 $\pm$ 475 & 3043 $\pm$ 741 \\
Tables referenced & 2.3 $\pm$ 0.8 & 5.3 $\pm$ 1.2\\
Joins per query & 1.9 $\pm$ 1.4 & 14.6 $\pm$ 5.8 \\
Logical conditions & 5.8 $\pm$ 5.9 & 32.2 $\pm$ 11.4 \\
Queries with aggregation functions (\%) & 100.0~\% & 98.1~\% \\
Queries with date/time operations (\%) & 78.3~\% & 100.0~\% \\
Queries with subqueries (\%) & 84.3~\% & 100.0~\% \\
Samples with temporal constraints (\%) & 52.2\% & 100.0\% \\
\hline
\hline
\end{tabular}
\vspace{5pt} 
\caption{Detailed statistics of the EpiAskKB and EpiCohoKB datasets. Global Statistics represent dataset-wide totals. Text Characteristics, Criteria Complexity, and SQL Complexity show averages per sample with standard deviations. Percentages indicate the proportion of samples with the specified characteristic. N/A denotes metrics not applicable to the dataset type.}
\label{tab:dataset-statistics}
\end{table}

We create two complementary knowledge bases (KBs) for our retrieval-augmented approach: \textit{EpiAskKB} and \textit{EpiCohoKB}. Table~\ref{tab:dataset-statistics} provides detailed statistics of both knowledge bases, highlighting their complementary nature and complexity.

EpiAskKB consists of 115 question-SQL pairs that capture typical analytical questions in observational studies, extending and refining the dataset from Ziletti and D'Ambrosi~\cite{ziletti-dambrosi-2024-retrieval} to better reflect common epidemiological research questions.

EpiCohoKB contains inclusion/exclusion criteria for observational studies, which we create through structured language model prompting. The generation integrates three key elements: (1) a prompt specifying OMOP CDM vocabulary usage and structural constraints for cohort criteria, (2) standard guidelines for study criteria (including temporal relationships, assessment windows, and observable time definitions) \cite{wang2021start}, and (3) real observational study protocols from the European Medicines Agency (EMA) catalog~\cite{hmaema2024catalogues} (89 samples) as exemplars for in-context learning.
This ensures both structural rigor and clinical authenticity.

EpiCohoKB covers diverse therapeutic areas and study types, including drug safety analyses, comparative effectiveness research, and healthcare utilization studies, creating a comprehensive benchmark for clinical criteria translation.
We generate an initial set of SQL queries through large language model (LLM) prompting, manually review and validate them for Snowflake SQL compatibility, resulting in 108 samples.
Data scientists with extensive expertise in epidemiological research and SQL query development for observational studies curated and validated both knowledge bases, ensuring both technical accuracy and clinical relevance of the final dataset.

\section{Methods} \label{sec:methodology}
Our approach translates clinical cohort definitions into executable SQL queries through a multi-stage pipeline (Fig.~\ref{fig:figure-workflow}). First, we use an LLM (Claude 3.5 Sonnet) to parse input criteria into a semi-structured format, explicitly identifying inclusion criteria, exclusion criteria, and index date definition. We opted for this approach over passing raw text directly to the final generation model for two primary reasons. First, this decomposition allows our system to apply RAG strategies at a more granular level (e.g., per-criterion retrieval in RAG+A and RAG+A+C). Second, it simplifies the complex task for the final SQL-generating LLM, allowing it to focus solely on translating structured logic into code, which we found empirically to reduce errors related to missed or incorrectly combined criteria.  We then compile an instruction prompt incorporating domain knowledge about OMOP CDM and SQL query construction, forming our baseline zero-shot (ZS) approach.
We implement three retrieval augmentation strategies: 
\begin{enumerate}
\item Criteria-level RAG (RAG + A) retrieves top-5 similar question-SQL pairs from EpiAskKB per criterion;  
\item Cohort-level RAG (RAG + C) retrieves top-5 similar cohort definitions from EpiCohoKB;
\item Combined RAG (RAG + A + C) performs both criteria-level and cohort-level retrieval. 
\end{enumerate}
This modular design enables query pattern reuse at different granularities - individual criteria can be composed into complex cohort definitions while preserving temporal relationships.
The augmented prompt is passed to the LLM to generate SQL with placeholder entities (e.g., [condition@hypertension]). This approach disentangles the query logic generation from the subsequent medical coding step where placeholders are mapped to specific concept IDs~\cite{ziletti-dambrosi-2024-retrieval}. The system implements a self-healing mechanism, which attempts to fix non-executable queries by feeding compiler errors back to the LLM for up to three iterations~\cite{pourreza-2023-dinsql}. The output is a cohort table with patient IDs and index dates.

To enhance interpretability for clinical users and enable criteria refinement, our system implements a patient funnel calculation pipeline (Fig.~\ref{fig:figure-workflow}h). While the primary output of our system is the final patient cohort (a list of patient IDs and index dates), the funnel provides a transparent, step-by-step view of cohort attrition by quantifying the remaining patient count as each rule is applied. For example, a user can see that an initial pool of $10\,000$ patients is reduced to $8\,000$ after meeting an age requirement, and further reduced to $5\,000$ after excluding those with a recent medication. This breakdown allows researchers to validate the logic of individual criteria and identify if a specific rule is unexpectedly strict or lenient. To implement this, the LLM generates separate SQL queries for each criterion. These are executed independently and combined sequentially using set operations (set intersections for inclusion criteria and set differences for exclusion criteria), tracking the remaining patient count at each stage.

\section{Evaluation} \label{sec:evaluation}
\subsection{Experimental setup} \label{sec:exp-setup}
We evaluated several state-of-the-art LLMs: Claude 3.5 Sonnet (hereafter \textit{Claude})~\cite{anthropic2024claude}, Gemini 2.0 Flash (\textit{Gemini})~\cite{geminiteam2023gemini}, LLAMA 3.1 70B~\cite{grattafiori-llama3-2024}, and GPT-4o~\cite{openai2023gpt4}. We also include reasoning models: o1~\cite{o1-openai2024learning} and DeepSeek R1 (\textit{deepseek})~\cite{deepseekai2025-deepseekr1}. 
All experiments were conducted in February 2025. We used a temperature of 0.0 for all models; top\_p was set to 0.9 for LLAMA 3.1 70B and the provider's default for all other models. Due to computational cost constraints, we adopted a two-stage evaluation strategy for o1, DeepSeek R1, and LLAMA 3.1 70B, testing them only in the zero-shot (ZS) setting and with our best-performing RAG + A + C configuration to establish a performance baseline and measure the gains from our optimal augmentation strategy.
% computational considerations
Input prompts to LLMs range from $1\,000$ to $3\,000$ tokens depending on the RAG strategy. The total end-to-end time for the full pipeline, including all generation steps and subsequent query executions on Snowflake, was typically 2 to 6 minutes, confirming its viability for near-interactive analysis.

Our pipeline employs a two-stage approach to handle medical entities. For the RAG step, we use entity masking to improve the relevance of retrieved examples~\cite{ziletti-dambrosi-2024-retrieval}. By replacing specific terms with generic labels (e.g., \textbf{endometriosis} $\rightarrow$ \textbf{CONDITION}), the system focuses on retrieving examples with similar analytical structures, rather than just similar medical concepts. This retrieval is powered by the \textsc{bge-large-en-v1.5} embedding model~\cite{zhang-etal-2023-retrieve}.
For the subsequent SQL generation, we instruct the LLM to use placeholders (e.g., condition@hypertension). This strategy decouples generating logical SQL from the specialized task of medical coding~\cite{limsopatham-collier-2016-normalising,portelli-etal-2022-generalizing,ziletti-etal-2022-medical,zhang-etal-2022-knowledge}. These placeholders are then resolved through a dedicated medical entity normalization process~\cite{ziletti-dambrosi-2024-retrieval}, which maps the natural language entity to a standardized OMOP Concept ID. This step is crucial for preventing the LLM from hallucinating invalid concept IDs and ensures the final query is both executable and accurate. This normalization also uses the \textsc{bge-large-en-v1.5} embeddings, with an additional verification layer from GPT-4o to ensure high fidelity~\cite{ziletti-dambrosi-2024-retrieval}.

The evaluation framework distinguishes between two key components: (1) the evaluation dataset consisting of inclusion/exclusion criteria paired with reference SQL queries, and (2) the healthcare database where these queries are executed. For the evaluation dataset, we use the 108 inclusion/exclusion criteria from EpiCohoKB described in Section \ref{sec:dataset}, employing a leave-one-out approach to prevent data leakage during retrieval. For the database component, we execute both generated and reference SQL queries on Optum's de-identified Clinformatics\textsuperscript{\textregistered} Data Mart Database (a major US claims database) converted to OMOP CDM to retrieve actual patient cohorts. Our methodology is compatible with any OMOP-formatted database, including the OMOP network (over 2.1 billion patient records worldwide~\cite{reich-2024-ohdsi}) and the freely available DE-SynPUF synthetic dataset~\cite{synpuf-cms-2008}.
For successful queries, we computed patient-level metrics (F1, precision, recall) and normalized cohort size differences based on matching patients, with temporal alignment assessed through exact matches and within a $\pm$~30-day window. Failed queries (invalid SQL or empty results) contribute zero scores to the final cross-sample averages.

\subsection{Experimental results} \label{sec:exp-results}
Results are shown in Table~\ref{tab:performance}, and outlined below.

\begin{sidewaystable}
\setlength{\tabcolsep}{4pt}
\renewcommand{\arraystretch}{1.2}  
\small
\centering
\begin{tabular}{l l | c c | c c c c | c c}
\hline
\hline
\multirow{2}{*}{Model name} & \multirow{2}{*}{Settings} & 
\multicolumn{2}{c|}{General metrics} & 
\multicolumn{4}{c|}{Patient-level metrics} & 
\multicolumn{2}{c}{Date-level metrics} \\
\cline{3-10}
& & Valid SQL & Retrieved & F1 & Prec. & Recall & Size sim. & Date overlap & Within 30d \\
\hline
\midrule
Claude 3.5 Sonnet & ZS & \textbf{100.0} & 78.1 & 57.8 & 61.0 & 63.1 & 62.8 & 75.9 & 74.9 \\
Claude 3.5 Sonnet & RAG+A & \underline{99.0} & 82.6 & 58.4 & 59.5 & 68.6 & 62.6 & 80.8 & 77.5 \\
Claude 3.5 Sonnet & RAG+C & \underline{99.0} & \underline{91.4} & 69.5 & 72.0 & 74.7 & 74.5 & 88.2 & 87.4 \\
Claude 3.5 Sonnet & RAG+A+C & \underline{99.0} & \textbf{93.3} & \underline{72.8} & \underline{75.6} & \underline{79.0} & \underline{77.6} & \textbf{91.7} & \textbf{90.5} \\
\midrule
Gemini 2.0 Flash & ZS & 96.2 & 55.2 & 42.4 & 43.3 & 44.5 & 46.6 & 51.7 & 50.9 \\
Gemini 2.0 Flash & RAG+A & \underline{99.0} & 62.9 & 41.6 & 42.1 & 48.6 & 46.0 & 58.7 & 58.4 \\
Gemini 2.0 Flash & RAG+C & \underline{99.0} & 89.5 & 68.0 & 68.9 & 75.2 & 73.5 & 85.4 & 84.6 \\
Gemini 2.0 Flash & RAG+A+C & \underline{99.0} & 91.3 & \textbf{75.4} & \textbf{76.6} & \textbf{79.7} & \textbf{79.8} & \underline{89.1} & \underline{89.1} \\
\midrule
GPT-4o & ZS & 84.8 & 61.0 & 35.4 & 38.1 & 39.7 & 42.1 & 52.9 & 52.5 \\
GPT-4o & RAG+A & 86.7 & 50.5 & 31.7 & 32.8 & 34.7 & 37.4 & 44.9 & 44.6 \\
GPT-4o & RAG+C & 97.1 & 91.4 & 72.1 & 73.1 & 76.8 & 77.7 & 87.0 & 87.9 \\
GPT-4o & RAG+A+C & 98.1 & 84.8 & 65.7 & 67.7 & 71.1 & 70.5 & 80.9 & 82.3 \\
\midrule
o1 & ZS & 9.5 & 5.7 & 4.3 & 5.2 & 4.2 & 4.5 & 5.4 & 5.0 \\
o1 & RAG+A+C & 78.1 & 73.3 & 60.0 & 62.0 & 62.2 & 63.0 & 71.5 & 70.0 \\
\midrule
DeepSeek R1 & ZS & 35.2 & 28.6 & 22.2 & 22.6 & 23.1 & 23.6 & 25.9 & 24.8  \\
DeepSeek R1 & RAG+A+C & 36.2 &31.4 & 23.6 & 25.0 & 25.4 & 25.4 & 29.6 & 30.2 \\
\midrule
LLAMA 3.1 70B & ZS & 37.1 & 17.1 & 8.5 & 9.6 & 9.0 & 11.0 & 16.2 & 14.5 \\
LLAMA 3.1 70B & RAG+A+C & 79.1 & 69.5 & 47.0 & 51.2 & 52.0 & 52.7 & 66.6 & 65.3 \\
\hline
\hline
\end{tabular}
\caption{Performance evaluation of text-to-SQL generation for patient cohort identification. \textit{Valid SQL} indicates syntactically correct queries, \textit{Retrieved} indicates queries that successfully retrieved patient data. Patient-level metrics evaluate cohort membership accuracy, while date-level metrics assess the temporal alignment of cohort index dates. Higher values are better, all values in percentage. Bold indicates best results, underlined shows second best.}
\label{tab:performance}
\end{sidewaystable}
\subsection*{Generated queries have high syntactic validity.} 
Most models show high SQL compilation rates with RAG augmentation, while also improving data retrieval success. The retrieved data rate increases substantially with RAG.
\subsection*{RAG strongly improves zero-shot performance.}
The combined RAG strategy (RAG + A + C) substantially outperforms ZS across all models. 
RAG + A + C significantly improves patient cohort identification across all metrics. Precision increases (Claude: $\uparrow$14.6pp, Gemini $\uparrow$33.3pp, GPT-4o $\uparrow$29.6pp), while maintaining high recall (>70\%).  Similar patterns emerge in temporal alignment.
The balanced improvement in precision and recall suggests that RAG helps models better understand both inclusion and exclusion criteria.
\subsection*{Complementary RAG strategies enhance performance.}
Cohort-level examples (RAG + C) are particularly effective. When compared to criteria-level RAG (RAG + A), cohort-level examples lead to substantially higher F1 scores. The combination of both strategies (RAG + A + C) further improves performance for most models, suggesting complementary benefits from both example types. Interestingly, GPT-4o shows slightly decreased performance with RAG + A + C compared to RAG + C ($\downarrow$6.4pp), indicating that longer input contexts may impact its generation capabilities. 
\subsection*{Claude and Gemini emerge as top performers.}
Claude and Gemini achieve comparable performance when augmented with RAG+A+C. Claude exhibits the strongest ZS performance, while Gemini shows excellent in-context learning capabilities ($\uparrow$33.3pp from ZS to RAG + A + C). Reasoning-focused models (o1, DeepSeek R1) struggle with SQL formatting despite their analytical capabilities.

To evaluate the patient funnel creation, for the best performing model (Claude RAG + A + C), we measure a 90\% normalized cohort size similarity compared to the single-query approach, validating the funnel decomposition strategy.

\subsection*{Error analysis}
To identify key areas for future improvement, we conduct a qualitative error analysis on the incorrect queries generated by our best-performing configuration (Claude 3.5 Sonnet with RAG+A+C).

The most significant challenges relate to the interpretation of complex temporal logic, a cornerstone of cohort study design. A frequent issue is the model's difficulty in correctly anchoring inclusion or exclusion criteria to the patient-specific index date. For instance, when defining a lookback window to identify new users of a drug, a query might incorrectly include the index date itself (e.g., using \texttt{BETWEEN DATEADD(day, -365, index\_date) AND index\_date}). This subtle discrepancy in boundary logic causes the index prescription to be found in the prior history, erroneously excluding every valid patient. Another common temporal challenge is atemporal exclusion; for a criterion to exclude patients with a diagnosis \emph{before} the index date, the model generates logic to exclude patients with that diagnosis at \emph{any time} in their record, incorrectly removing those who develop the condition after the index event.

Other significant error patterns also emerge. We observe challenges in understanding the specific conventions of the OMOP-CDM schema. For example, a query for ``psychiatric evaluation'' might incorrectly search the \texttt{VISIT\_OCCURRENCE} table instead of the correct \texttt{PROCEDURE\_OCCURRENCE} table. Similarly, a query for ``mental health visits'' often omits the required join from the \texttt{VISIT\_OCCURRENCE} table to the \texttt{PROVIDER} table to filter by specialty. We also identify errors in complex query construction. A key example involves suboptimal sequences of operations, such as applying a date-range filter before an aggregation intended to find a patient's first-ever event. This leads to the model identifying the first event \textit{within the specified window}, rather than the true \textit{first-ever event} for that patient, thereby misidentifying the index date. Finally, some errors stem from a less nuanced interpretation of clinical concepts, such as translating ``chronic use'' into a check for any single prescription in the \texttt{DRUG\_EXPOSURE} table instead of the more appropriate logic of identifying a long-term drug era in the \texttt{DRUG\_ERA} table.

Collectively, these findings indicate that the primary opportunities for advancement lie not in improving syntactic SQL generation, but in enhancing the model's reasoning about the complex temporal relationships, data model conventions, and clinical semantics.

\section{Related Work} \label{sec:Related}
\subsection*{Cohort generation from EHR data.}
Rule-based systems can achieve high precision but require substantial expert curation~\cite{banda2017electronic,hripcsak2019facilitating,ohdsi2021book,privitera2024phenex}. While early work combined rules and machine learning~\cite{yuan-2019-criteria}, recent approaches leverage LLMs: for OMOP-CDM clinical trial queries~\cite{park2024-criteria}, for structured criteria extraction~\cite{melnichenko2023designing}, for cross-schema templated queries~\cite{dobbins-et-al-2023-leafai}, for phenotyping~\cite{yan-et-al-2024-large}, and for eligibility criteria structuring~\cite{wong-et-al-2023-scaling}. 
Several EHR text-to-SQL datasets exist~\cite{wang-etal-2020-text,raghavan-etal-2021-emrkbqa,lee-2022-ehrsql,tarbell2023-towards,ziletti-dambrosi-2024-retrieval}, though none maps observational study criteria to OMOP-SQL.
Ziletti and D'Ambrosi~\cite{ziletti-dambrosi-2024-retrieval} previously introduced RAG-enhanced epidemiological text-to-SQL for question answering. This work extends that approach to the more complex task of cohort generation, which requires preserving temporal relationships, handling logical combinations of criteria, and generating patient-level identifiers with index dates. As evidenced by the statistics in Table~\ref{tab:dataset-statistics}, cohort generation involves significantly more complex queries with more joins, logical conditions, and temporal constraints compared to analytical questions. Our innovations include two-level retrieval augmentation, patient funnel generation, and comprehensive evaluation of patient-level precision and recall with temporal alignment.

\subsection*{Text-to-SQL with in-context learning.}
Recent work shows LLMs excel at text-to-SQL tasks when enhanced with in-context learning~\cite{rajkumar2022evaluating}, particularly through optimized example selection~\cite{nan-etal-2023-enhancing,gao-2023-text}, query decomposition~\cite{pourreza-2023-dinsql}, diverse demonstrations~\cite{chang2023prompt,chang-fosler-lussier-2023-selective}, ensemble approaches~\cite{gao2025-preview-xiyansql}, and chain-of-thought prompting~\cite{zhang-etal-2023-act}.

\section{Conclusions} \label{sec:conclusion}
This work shows that LLMs with RAG can effectively identify patient cohorts from EHR data.
The pipeline combines medical concept standardization, criteria parsing, and two-level retrieval to handle complex temporal relationships, while the patient funnel provides interpretable intermediate metrics.
Future applications to clinical trial recruitment could help bridge the gap between observational and interventional research.

\section{Limitations} \label{sec:limitations}
Our work has several limitations. The modest size of our dataset may not capture the full complexity and variety of real-world observational studies. The current system lacks formal uncertainty estimates for its predictions, making it difficult to identify cases where human review is particularly important. Additionally, the accuracy of intermediate preprocessing steps like semi-structured parsing and entity normalization was not separately evaluated, which would be required to fully characterize the pipeline's robustness. Although the patient funnel offers valuable transparency into the cohort attrition process, the generated SQL itself can be difficult to interpret for users unfamiliar with the language. This difficulty may limit the ability of clinical researchers to fully validate the query's underlying logic. The current methodology relies on OMOP-CDM, and although it could be in principle adapted to other data models through in-context learning, this was not tested.

\bibliographystyle{splncs04}
\bibliography{anthology,custom-short}

\begin{thebibliography}{10}
\providecommand{\url}[1]{\texttt{#1}}
\providecommand{\urlprefix}{URL }
\providecommand{\doi}[1]{https://doi.org/#1}

\bibitem{anthropic2024claude}
{Anthropic AI}: Model card for claude 3 models.
  \url{https://docs.anthropic.com/en/docs/resources/model-card} (2024),
  accessed: February 15, 2024

\bibitem{banda2017electronic}
Banda, J.M., Halpern, Y., Sontag, D., Shah, N.H.: Electronic phenotyping with
  {APHRODITE} and the {Observational Health Sciences and Informatics (OHDSI)}
  data network. AMIA Joint Summits on Translational Science proceedings
  \textbf{2017},  48--57 (2017)

\bibitem{casey-et-al-2016-using}
Casey, J.A., Schwartz, B.S., Stewart, W.F., Adler, N.E.: Using electronic
  health records for population health research: A review of methods and
  applications. Annual Review of Public Health  \textbf{37}(Volume 37, 2016),
  61--81 (2016).
  \doi{https://doi.org/10.1146/annurev-publhealth-032315-021353},
  \url{https://www.annualreviews.org/content/journals/10.1146/annurev-publhealth-032315-021353}

\bibitem{chang2023prompt}
Chang, S., Fosler-Lussier, E.: How to prompt llms for text-to-sql: A study in
  zero-shot, single-domain, and cross-domain settings (2023)

\bibitem{chang-fosler-lussier-2023-selective}
Chang, S., Fosler-Lussier, E.: Selective demonstrations for cross-domain
  text-to-{SQL}. In: Bouamor, H., Pino, J., Bali, K. (eds.) Findings of the
  Association for Computational Linguistics: EMNLP 2023. pp. 14174--14189.
  Association for Computational Linguistics, Singapore (Dec 2023).
  \doi{10.18653/v1/2023.findings-emnlp.944},
  \url{https://aclanthology.org/2023.findings-emnlp.944}

\bibitem{dobbins-et-al-2023-leafai}
Dobbins, N.J., Han, B., Zhou, W., Lan, K.F., Kim, H.N., Harrington, R., Uzuner,
  O., Yetisgen, M.: Leafai: query generator for clinical cohort discovery
  rivaling a human programmer. Journal of the American Medical Informatics
  Association  \textbf{30}(12),  1954--1964 (08 2023).
  \doi{10.1093/jamia/ocad149}, \url{https://doi.org/10.1093/jamia/ocad149}

\bibitem{hmaema2024catalogues}
{EMA}: Hma-ema catalogues of real-world data sources and studies (2024),
  \url{https://catalogues.ema.europa.eu/}, accessed: \today

\bibitem{gao-2023-text}
Gao, D., Wang, H., Li, Y., Sun, X., Qian, Y., Ding, B., Zhou, J.: Text-to-sql
  empowered by large language models: A benchmark evaluation (2023)

\bibitem{gao2025-preview-xiyansql}
Gao, Y., Liu, Y., Li, X., Shi, X., Zhu, Y., Wang, Y., Li, S., Li, W., Hong, Y.,
  Luo, Z., Gao, J., Mou, L., Li, Y.: A preview of xiyan-sql: A multi-generator
  ensemble framework for text-to-sql (2025),
  \url{https://arxiv.org/abs/2411.08599}

\bibitem{geminiteam2023gemini}
{Gemini Team}: Gemini: A family of highly capable multimodal models. Tech.
  rep., Google (12 2023),
  \url{https://storage.googleapis.com/deepmind-media/gemini_1_report.pdf},
  accessed: February 15, 2024

\bibitem{grattafiori-llama3-2024}
Grattafiori, A., et~al.: The llama 3 herd of models (2024),
  \url{https://arxiv.org/abs/2407.21783}

\bibitem{deepseekai2025-deepseekr1}
Guo, D., et~al.: Deepseek-r1: Incentivizing reasoning capability in llms via
  reinforcement learning (2025), \url{https://arxiv.org/abs/2501.12948}

\bibitem{hripcsak2019facilitating}
Hripcsak, G., Shang, N., Peissig, P.L., Rasmussen, L.V., Liu, C., Benoit, B.,
  Carroll, R.J., Carrell, D.S., Denny, J.C., Dikilitas, O., Gainer, V.S.,
  Howell, K.M., Klann, J.G., Kullo, I.J., Lingren, T., Mentch, F.D., Murphy,
  S.N., Natarajan, K., Pacheco, J.A., Wei, W.Q., Wiley, K., Weng, C.:
  Facilitating phenotype transfer using a common data model. Journal of
  Biomedical Informatics  \textbf{96},  103253 (2019).
  \doi{https://doi.org/10.1016/j.jbi.2019.103253},
  \url{https://www.sciencedirect.com/science/article/pii/S1532046419301728}

\bibitem{lee-2022-ehrsql}
Lee, G., Hwang, H., Bae, S., Kwon, Y., Shin, W., Yang, S., Seo, M., Kim, J.Y.,
  Choi, E.: Ehrsql: A practical text-to-sql benchmark for electronic health
  records. In: Koyejo, S., Mohamed, S., Agarwal, A., Belgrave, D., Cho, K., Oh,
  A. (eds.) Advances in Neural Information Processing Systems. vol.~35, pp.
  15589--15601. Curran Associates, Inc. (2022)

\bibitem{limsopatham-collier-2016-normalising}
Limsopatham, N., Collier, N.: Normalising medical concepts in social media
  texts by learning semantic representation. In: Erk, K., Smith, N.A. (eds.)
  Proceedings of the 54th Annual Meeting of the Association for Computational
  Linguistics (Volume 1: Long Papers). pp. 1014--1023. Association for
  Computational Linguistics, Berlin, Germany (Aug 2016).
  \doi{10.18653/v1/P16-1096}, \url{https://aclanthology.org/P16-1096}

\bibitem{melnichenko2023designing}
Melnichenko, O.: Designing multi-step cohort creation flow for biomedical
  datasets. Microsoft Technical Community Blog  (2023),
  \url{https://techcommunity.microsoft.com/blog/healthcareandlifesciencesblog/designing-multi-step-cohort-creation-flow-for-biomedical-datasets/3964454}

\bibitem{nan-etal-2023-enhancing}
Nan, L., Zhao, Y., Zou, W., Ri, N., Tae, J., Zhang, E., Cohan, A., Radev, D.:
  Enhancing text-to-{SQL} capabilities of large language models: A study on
  prompt design strategies. In: Bouamor, H., Pino, J., Bali, K. (eds.) Findings
  of the Association for Computational Linguistics: EMNLP 2023. pp.
  14935--14956. Association for Computational Linguistics, Singapore (Dec
  2023). \doi{10.18653/v1/2023.findings-emnlp.996},
  \url{https://aclanthology.org/2023.findings-emnlp.996}

\bibitem{ohdsi2021book}
{OHDSI Collaborative}: The book of {OHDSI}. In: The Book of {OHDSI}, chap.~10.
  Observational Health Data Sciences and Informatics (2021),
  \url{http://book.ohdsi.org}

\bibitem{omop-cdm-2023}
OMOP-CDM: Omop cdm common data model.
  \url{https://ohdsi.github.io/CommonDataModel/} (2023), accessed: January 23,
  2024

\bibitem{openai2023gpt4}
OpenAI: Gpt-4 technical report (2023)

\bibitem{o1-openai2024learning}
{OpenAI}: Learning to reason with llms (September 2024),
  \url{https://openai.com/index/learning-to-reason-with-llms/}, accessed:
  September 2024

\bibitem{park2024-criteria}
Park, J., Fang, Y., Ta, C., Zhang, G., Idnay, B., Chen, F., Feng, D., Shyu, R.,
  Gordon, E.R., Spotnitz, M., Weng, C.: Criteria2query 3.0: Leveraging
  generative large language models for clinical trial eligibility query
  generation. Journal of Biomedical Informatics  \textbf{154},  104649 (2024).
  \doi{https://doi.org/10.1016/j.jbi.2024.104649},
  \url{https://www.sciencedirect.com/science/article/pii/S1532046424000674}

\bibitem{portelli-etal-2022-generalizing}
Portelli, B., Scaboro, S., Santus, E., Sedghamiz, H., Chersoni, E., Serra, G.:
  Generalizing over long tail concepts for medical term normalization. In:
  Goldberg, Y., Kozareva, Z., Zhang, Y. (eds.) Proceedings of the 2022
  Conference on Empirical Methods in Natural Language Processing. pp.
  8580--8591. Association for Computational Linguistics, Abu Dhabi, United Arab
  Emirates (Dec 2022). \doi{10.18653/v1/2022.emnlp-main.588},
  \url{https://aclanthology.org/2022.emnlp-main.588}

\bibitem{pourreza-2023-dinsql}
Pourreza, M., Rafiei, D.: {DIN}-{SQL}: Decomposed in-context learning of
  text-to-{SQL} with self-correction. In: Thirty-seventh Conference on Neural
  Information Processing Systems (2023),
  \url{https://openreview.net/forum?id=p53QDxSIc5}

\bibitem{privitera2024phenex}
Privitera, S., Hartenstein, A.: Phenex: Automatic phenotype extractor.
  \url{https://github.com/Bayer-Group/PhenEx} (2024), accessed: \today

\bibitem{raghavan-etal-2021-emrkbqa}
Raghavan, P., Liang, J.J., Mahajan, D., Chandra, R., Szolovits, P.: emr{KBQA}:
  A clinical knowledge-base question answering dataset. In: Demner-Fushman, D.,
  Cohen, K.B., Ananiadou, S., Tsujii, J. (eds.) Proceedings of the 20th
  Workshop on Biomedical Language Processing. pp. 64--73. Association for
  Computational Linguistics, Online (Jun 2021).
  \doi{10.18653/v1/2021.bionlp-1.7},
  \url{https://aclanthology.org/2021.bionlp-1.7}

\bibitem{rajkumar2022evaluating}
Rajkumar, N., Li, R., Bahdanau, D.: Evaluating the text-to-sql capabilities of
  large language models. ArXiv  \textbf{abs/2204.00498} (2022),
  \url{https://api.semanticscholar.org/CorpusID:247922681}

\bibitem{reich-2024-ohdsi}
Reich, C., Ostropolets, A., Ryan, P., Rijnbeek, P., Schuemie, M., Davydov, A.,
  Dymshyts, D., Hripcsak, G.: {OHDSI Standardized Vocabularies—a large-scale
  centralized reference ontology for international data harmonization}. Journal
  of the American Medical Informatics Association p. ocad247 (01 2024).
  \doi{10.1093/jamia/ocad247}, \url{https://doi.org/10.1093/jamia/ocad247}

\bibitem{sherman-et-al-2016-real}
Sherman, R.E., Anderson, S.A., Pan, G.J.D., Gray, G.W., Gross, T., Hunter,
  N.L., LaVange, L., Marinac-Dabic, D., Marks, P.W., Robb, M.A., Shuren, J.,
  Temple, R., Woodcock, J., Yue, L.Q., Califf, R.M.: Real-world evidence —
  what is it and what can it tell us? New England Journal of Medicine
  \textbf{375}(23),  2293--2297 (2016). \doi{10.1056/NEJMsb1609216},
  \url{https://www.nejm.org/doi/full/10.1056/NEJMsb1609216}

\bibitem{synpuf-cms-2008}
SynPUF: Medicare claims synthetic public use files (synpufs).
  \url{https://www.cms.gov/data-research/statistics-trends-and-reports/medicare-claims-synthetic-public-use-files}
  (2010), accessed: January 23, 2024

\bibitem{tarbell2023-towards}
Tarbell, R., Choo, K.K.R., Dietrich, G., Rios, A.: Towards understanding the
  generalization of medical text-to-sql models and datasets. AMIA ... Annual
  Symposium proceedings. AMIA Symposium  \textbf{2023},  669--678 (2023),
  \url{https://api.semanticscholar.org/CorpusID:257687830}

\bibitem{fda2024realworld}
{U.S. FDA}: Real-world data: Assessing electronic health records and medical
  claims data to support regulatory decision-making for drug and biological
  products. Guidance for Industry FDA-2020-D-2307, U.S. Food and Drug
  Administration (July 2024),
  \url{https://www.fda.gov/regulatory-information/search-fda-guidance-documents/real-world-data-assessing-electronic-health-records-and-medical-claims-data-support-regulatory},
  final Level 1 Guidance

\bibitem{wang-etal-2020-text}
Wang, P., Shi, T., Reddy, C.K.: Text-to-sql generation for question answering
  on electronic medical records. In: Proceedings of The Web Conference 2020. p.
  350–361. WWW '20, Association for Computing Machinery, New York, NY, USA
  (2020). \doi{10.1145/3366423.3380120},
  \url{https://doi.org/10.1145/3366423.3380120}

\bibitem{wang2021start}
Wang, S.V., Pinheiro, S., Hua, W., Arlett, P., Uyama, Y., Berlin, J.A.,
  Bartels, D.B., Kahler, K.H., Bessette, L.G., Schneeweiss, S.: Start-rwe:
  structured template for planning and reporting on the implementation of real
  world evidence studies. BMJ  \textbf{372} (2021). \doi{10.1136/bmj.m4856},
  \url{https://www.bmj.com/content/372/bmj.m4856}

\bibitem{wong-et-al-2023-scaling}
Wong, C., Zhang, S., Gu, Y., Moung, C., Abel, J., Usuyama, N., Weerasinghe, R.,
  Piening, B., Naumann, T., Bifulco, C., Poon, H.: Scaling clinical trial
  matching using large language models: A case study in oncology. In:
  Deshpande, K., Fiterau, M., Joshi, S., Lipton, Z., Ranganath, R., Urteaga,
  I., Yeung, S. (eds.) Proceedings of the 8th Machine Learning for Healthcare
  Conference. Proceedings of Machine Learning Research, vol.~219, pp. 846--862.
  PMLR (11--12 Aug 2023), \url{https://proceedings.mlr.press/v219/wong23a.html}

\bibitem{yan-et-al-2024-large}
Yan, C., Ong, H.H., Grabowska, M.E., Krantz, M.S., Su, W.C., Dickson, A.L.,
  Peterson, J.F., Feng, Q., Roden, D.M., Stein, C.M., Kerchberger, V.E., Malin,
  B.A., Wei, W.Q.: Large language models facilitate the generation of
  electronic health record phenotyping algorithms. Journal of the American
  Medical Informatics Association  \textbf{31}(9),  1994--2001 (04 2024).
  \doi{10.1093/jamia/ocae072}, \url{https://doi.org/10.1093/jamia/ocae072}

\bibitem{yuan-2019-criteria}
Yuan, C., Ryan, P.B., Ta, C., Guo, Y., Li, Z., Hardin, J., Makadia, R., Jin,
  P., Shang, N., Kang, T., Weng, C.: {Criteria2Query: a natural language
  interface to clinical databases for cohort definition}. Journal of the
  American Medical Informatics Association  \textbf{26}(4),  294--305 (02
  2019). \doi{10.1093/jamia/ocy178}, \url{https://doi.org/10.1093/jamia/ocy178}

\bibitem{zhang-etal-2023-act}
Zhang, H., Cao, R., Chen, L., Xu, H., Yu, K.: {ACT}-{SQL}: In-context learning
  for text-to-{SQL} with automatically-generated chain-of-thought. In: Bouamor,
  H., Pino, J., Bali, K. (eds.) Findings of the Association for Computational
  Linguistics: EMNLP 2023. pp. 3501--3532. Association for Computational
  Linguistics, Singapore (Dec 2023). \doi{10.18653/v1/2023.findings-emnlp.227},
  \url{https://aclanthology.org/2023.findings-emnlp.227}

\bibitem{zhang-etal-2023-retrieve}
Zhang, P., Xiao, S., Liu, Z., Dou, Z., Nie, J.Y.: Retrieve anything to augment
  large language models (2023)

\bibitem{zhang-etal-2022-knowledge}
Zhang, S., Cheng, H., Vashishth, S., Wong, C., Xiao, J., Liu, X., Naumann, T.,
  Gao, J., Poon, H.: Knowledge-rich self-supervision for biomedical entity
  linking. In: Goldberg, Y., Kozareva, Z., Zhang, Y. (eds.) Findings of the
  Association for Computational Linguistics: EMNLP 2022. pp. 868--880.
  Association for Computational Linguistics, Abu Dhabi, United Arab Emirates
  (Dec 2022). \doi{10.18653/v1/2022.findings-emnlp.61},
  \url{https://aclanthology.org/2022.findings-emnlp.61}

\bibitem{ziletti-etal-2022-medical}
Ziletti, A., Akbik, A., Berns, C., Herold, T., Legler, M., Viell, M.: {M}edical
  coding with biomedical transformer ensembles and zero/few-shot learning. In:
  Loukina, A., Gangadharaiah, R., Min, B. (eds.) Proceedings of the 2022
  Conference of the North American Chapter of the Association for Computational
  Linguistics: Human Language Technologies: Industry Track. pp. 176--187.
  Association for Computational Linguistics, Hybrid: Seattle, Washington +
  Online (Jul 2022). \doi{10.18653/v1/2022.naacl-industry.21},
  \url{https://aclanthology.org/2022.naacl-industry.21}

\bibitem{ziletti-dambrosi-2024-retrieval}
Ziletti, A., {D'Ambrosi}, L.: Retrieval augmented text-to-{SQL} generation for
  epidemiological question answering using electronic health records. In:
  Naumann, T., Ben~Abacha, A., Bethard, S., Roberts, K., Bitterman, D. (eds.)
  Proceedings of the 6th Clinical Natural Language Processing Workshop. pp.
  47--53. Association for Computational Linguistics, Mexico City, Mexico (Jun
  2024). \doi{10.18653/v1/2024.clinicalnlp-1.4},
  \url{https://aclanthology.org/2024.clinicalnlp-1.4}

\end{thebibliography}

\end{document}